\documentclass{article}
\usepackage{graphicx}
\usepackage{amsmath}
\usepackage{arxiv}
\usepackage{hyperref}
\usepackage[utf8]{inputenc} % allow utf-8 input
\usepackage[T1]{fontenc}    % use 8-bit T1 fonts
\usepackage{hyperref}       % hyperlinks
\usepackage{url}            % simple URL typesetting
\usepackage{booktabs}       % professional-quality tables
\usepackage{amsfonts}       % blackboard math symbols
\usepackage{nicefrac}       % compact symbols for 1/2, etc.
\usepackage{microtype}      % microtypography
\usepackage{lipsum}

\usepackage{biblatex}

\title{Learning to Optimize Under Constraints with Unsupervised Deep Neural Networks}

\author{
Seyedrazieh Bayati\thanks{This paper is an extension of Seyedrazieh Bayati's Master's thesis \cite{bayati2020machine} in Electrical Engineering department of the University of British Columbia.} \\
Department of Electrical and Computer Engineering\\
  University of British Columbia V6T 1Z4\\
  Vancouver, BC Canada \\
  \texttt{razibayati20@gmail.com} \\
  \And
      Faramarz Jabbarvaziri\\
  Department of Electrical and Computer Engineering\\
  University of British Columbia V6T 1Z4\\
  Vancouver, BC Canada \\
  \texttt{jabbarva@ece.ubc.ca}

}

\begin{document}
\maketitle

\begin{abstract}
In this paper, we propose a machine learning (ML) method to learn how to solve a generic constrained continuous optimization problem. To the best of our knowledge, the generic methods that learn to optimize, focus on unconstrained optimization problems and those dealing with constrained problems are not easy-to-generalize. This approach is quite useful in optimization tasks where the problem's parameters constantly change, and require resolving the optimization task per parameter update. In such problems, the computational complexity of optimization algorithms such as gradient descent or interior point method preclude near-optimal designs in real-time applications. In this paper, we propose an unsupervised deep learning (DL) solution for solving constrained optimization problems in real-time by relegating the main computation load to offline training phase. This paper's main contribution is proposing a method for enforcing the equality and inequality constraints to the DL-generated solutions for generic optimization tasks.
\end{abstract}

% keywords can be removed

\section{Introduction}

% Although we can consider the parameters of an optimization problem as the input of the problem and the optimal point and value as the output, we often are interested in solving it for a set of constant parameters which is a one-time task. 

% Now assume that in an optimization problem the mapping from the parameters to optimal point and value forms a continuous function. If we have a constant set of parameters, there is no point in trying to find this function (mapping) and solving the optimization problem only once would be enough. 
In case the nature of the optimization problem requires constantly re-solving the optimization for different sets of input parameters, knowing a mapping from the set of parameters to the optimal solution would be extremely helpful. This way, instead of running an optimization algorithm online, to find the optimal solution for each input parameter, we merely need to feed the parameters to the mapping function and output the near-optimal solution with less computational complexity.

Consider the following unconstrained optimization problem
\begin{equation}
    \min_x \ f(x;p),
\end{equation}
where, $p$ a tensor containing the parameters of the problem. If $p$ varies quite often, instead of solving this problem numerically each time, we would rather find the mapping between the optimal point ($x^*$) and $p$. In such a case, a neural network can be trained in an unsupervised or supervised manner to find the mentioned mapping.

\section{Deep Learning-Based Optimization}
We design a deep neural network (DNN) as follows:
\begin{equation}
    x^* = \mathrm{DNN}(p).
\end{equation}
As suggested by \cite{johnson2016perceptual,8444648,sun2017learning}, an unsupervised learning method can potentially find the aforementioned mapping by setting the loss function of a DNN equal to the objective function of the optimization problem.
\begin{equation}
    \mathcal{L} = f(x;p)
\end{equation}

As an alternative, some employed reinforced learning (RL) to address similar problems. A 2016 study by Kie Li et al. used RL to solve unconstrained continuous optimization problems \cite{li2016learning} \footnote{This method is explained in \hyperlink{https://bair.berkeley.edu/blog/2017/09/12/learning-to-optimize-with-rl/}{https://bair.berkeley.edu/blog/2017/09/12/learning-to-optimize-with-rl/}.}. In addition, \cite{cappart2020combining,miagkikh1999approach} use RL to solve unconstrained discrete combinatorial optimization problems.

What has been missing in all of the mentioned works is the consideration of generic equality and inequality constraints. To the best of our knowledge, a method for learning to optimize an objective function under some generic constraints does not exist. To this end, \cite{Wei, 8586870,8847377,8935405} use DL for solving optimization tasks and to enforce their simple constraints they resort to the expert knowledge to alter the solution. The mentioned articles solve wireless network beamforming problems under power constraint, and they scale the beamforming vector by multiplying it with a normalizing factor or use saturating activation function in the output layer of the DNN to meet the power constraint of the norm of the beamforming vector. In this paper, we introduce a generic method for imposing any inequality or equality constraints.

\section{Piece-wise Regularization}

As mentioned before, we assume that $p$ is a tensor containing all parameters of the optimization problem (the problem setup) in a certain state and $\mathcal{P}$ is the set of many many $p$ tensors in various scenarios\footnote{By scenario we mean a certain problem setup.} (i.e., $p\in\mathcal{P}$). Also, let $x^*$ denote the optimum solution corresponding to $p$ and $\mathcal{X}^*$ such that $x^*\in \mathcal{X}^*$ denote the set of all optimum points corresponding to the set $\mathcal{P}$.

For each parameter tensor $p$ we want to generate $x^*$ through a DNN such that it minimizes the the objective function $f_0(x;p)$ under the set of the constraints $f_i(x;p) \le   c_i$ and $h_i(x;p) = b_i$. 
In this method, we set the loss function of the DNN equal to the objective function and add penalty terms to the objective function as follows:
\begin{equation}
\mathcal{L}(x^*;p) =  f_0(x^*;p) + \sum_i \mathbb{I}(f_i(x^*;p)- c_i )+ \sum_j \mathbb{I}(h_j(x^*;p)- b_j )
\end{equation}
where, $\mathbb{I}$ is defined as below
\begin{equation}
    \mathbb{I}(x) = \begin{cases}
    0\  &; \ x \leq 0 \\
    \infty \  &; \ x > 0
    \end{cases}
\end{equation}
In each epoch, our target would be to minimize the mean of the loss, i.e., $\overline{\mathcal{L}(\mathcal{X}^*;\mathcal{P})} =\frac{1}{|\mathcal{P}|}\sum_{k\in|\mathcal{P}|}\mathcal{L}(x_k^*;p_k)$. This way, we ensure that the constraints are met as the indicator function disposes of infeasible solutions by sending the value of the loss to infinity. Nonetheless, since the gradient of $\sum_i \mathbb{I}(f_i(x^*;p_k)- c_i )+ \sum_j \mathbb{I}(h_j(x^*;p_k)- b_j )$ is always zero, escaping the infeasible solutions can only happen by chance and the network cannot learn how to choose feasible solutions. 

A tweak for this issue would be to use penalty terms with non-zero gradient to penalize the infeasible solutions instead of using the $\mathbb{I}$ function. Our target is to ensure that the loss value outside of the feasible set becomes large enough such that no infeasible solution can minimize the $\mathcal{L}$. 

Let us define the "deviation from feasibility" for inequality constraints as follows.
\begin{equation}
    F_i(x^*;p) =  
    \begin{cases}
          0  \ & ;  f_i(x^*;p) \ \leq c_i \\
          \eta_i(f_i(x^*;p) - c_i)^\gamma \ & ; \ \mathrm{else}
    \end{cases}
\end{equation}
and for equality constraints 
\begin{equation}
    H_j(x^*;p) =  
    \begin{cases}
           0 \ &;  h_j(x^*;p) = b_j \\
          \eta_j|h_j(x^*;p) - b_j|^\gamma \ & ; \ \mathrm{else}
    \end{cases}.
\end{equation}
where, $\eta_i \ge 0$ are constant factor that tune the effect of the regulating terms, and $\gamma \geq 1$. Now we define the penalty term as follows
\begin{equation}
    \Omega(x^*;p) = \sum_i F_i(x^*;p) + \sum_j H_j(x^*;p),
\end{equation}

In this method, for a solution $x^*$, if none of the constraints get violated, $\Omega$ becomes $0$, and there is no penalty; otherwise, there is a penalty for each constraint that is violated proportional to the amount of violation (see Fig. \ref{penalty}). We define the mean loss as follows
\begin{equation}
\overline{\mathcal{L}(\mathcal{X}^*;\mathcal{P})}=\frac{1}{|\mathcal{P}|} \sum_{k\in\mathcal{P}}  (f_0(x_k^*;p_k)+ \Omega(x_k^*;p_k))
\end{equation}
\begin{figure}
    \centering
    \includegraphics[scale=0.65]{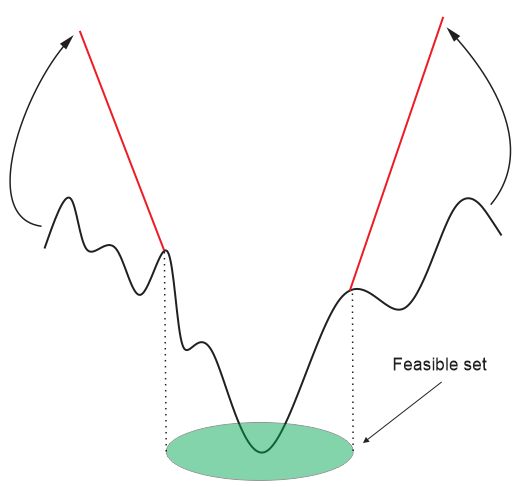}
    \caption{Penalizing solutions outside of the feasible set using piece-wise regularization method.}
    \label{penalty}
\end{figure} 

\section{Complexity Analysis}

The complexity of training a neural network that has $n$ inputs, $l$ hidden layers each one with $m_1,m_2,...,m_l$ neurons and $k$ outputs with back-propagation algorithm after $N_{\mathrm{e}}$ epochs and $N_{\mathrm{s}}$ samples is $\mathcal{O}(N_{\mathrm{e}}N_{\mathrm{s}}(nm_1+\sum_{i=1}^{l-1} m_i m_{i+1} + m_l k))$. However, in the forward path it is only $\mathcal{O}(nm_1+\sum_{i=1}^{l-1} m_i m_{i+1} + m_l k)$. 

The beauty of the proposed scheme is that it takes an enormous chunk of the computational complexity offline. This enables the forward path to deliver solution with a low computational complexity compared to online optimization algorithms such as Interior point method which is often used for non-convex optimizations problems. Interior point method has the worst-case computational complexity of $\mathcal{O}(\max\{n,m\}^4 \sqrt n \log(\frac{1}{\epsilon}))$, where $n$ is the number of variables, $m$ is the number of constraints, and $\epsilon$ is the solution accuracy \cite{interior}.

% \section{Impact of Perturbations}
% A sensitivity analysis on the results of DNN-based optimization approach and conventional optimization approaches will be provided here and the effect of SGD-based training method on producing robust DNN-based solutions will be discussed. \textbf{[to be done]}

% The reason we think it should work better than than the additive regularization is that the regularized objective function remains continuous in the border of feasible and infeasible solutions.
% \begin{figure}
% \centering
%   \includegraphics[width=.5\linewidth]{add.png}
%   \caption{Additive}
% \end{figure}

% \begin{figure}
% \centering
%   \includegraphics[width=.5\linewidth]{mul.png}
%   \caption{Multiplicative}
% \end{figure}

\section{Test Cases}
In this section, we test the proposed method on some test functions known as artificial landscapes \cite{test_functions} under some linear and non-linear constraints. In these tests we use the proposed piece-wise regularization with $\eta_i=10^8$ and $\gamma = 2$.
% \subsection{Convex constrained problem}
% Consider the following convex optimization problem:
% \begin{equation}
% \begin{cases}
%     \min_x (x-a)^2 \\
%     \mathrm{s.t.} \ x \leq 0.5
% \end{cases}
% \end{equation}
% The obvious optimal point of this problem is as follows 
% \begin{equation}
% x^* =
% \begin{cases}
%     a\ ;\  \mathrm{if} \ a\leq 0.5 \\
%     0.5 \ ;\  \mathrm{if} \ a\ge 0.5
% \end{cases}
% \end{equation}
% Using multiplicative regularization we were able to solve this problem. The constraint is also met.

% \textbf{Setup:}\\
% Samples: 1000 in range [-10,10]

% optimizer = torch.optim.Adam(model.parameters(), lr=.001)\\
% scheduler= torch.optim.lrScheduler.ReduceLROnPlateau(optimizer,mode=min,factor=0.8,patience=10,verbose=False)

\subsection{Rosenbrock's Function with One Constraint}
\begin{equation}
\begin{cases}
    \min_x c_1(x_2-x_1^2)^2 + (c_2-x_1)^2 \\
    \mathrm{s.t.} \ x_1^2+x_2^2 \leq 1
\end{cases}
\end{equation}
where, $c_1, c_2$ are the parameters of this problem. We employed a DNN with a structure $(D_{in}, H_{layer1}, H_{layer2}, D_{out}) = (2, 20, 20, 2 )$ and for training we used 1000 samples uniformly distributed in range $0\leq c_1 \leq 30$ and $0\leq c_2 \leq 1$. We used the adaptive moment estimation (ADAM) optimizer for training the DNN and got the results shown in Table~\ref{ros_table}. %Pytorch code of this example is available in \href{https://colab.research.google.com/drive/1rBhv_DyLV7Bie68izB9J3vB88xFyU9t9?usp=sharing}{[Colab: Ros-1-constraint]}.

\begin{figure}
    \centering
    \includegraphics[scale=0.4]{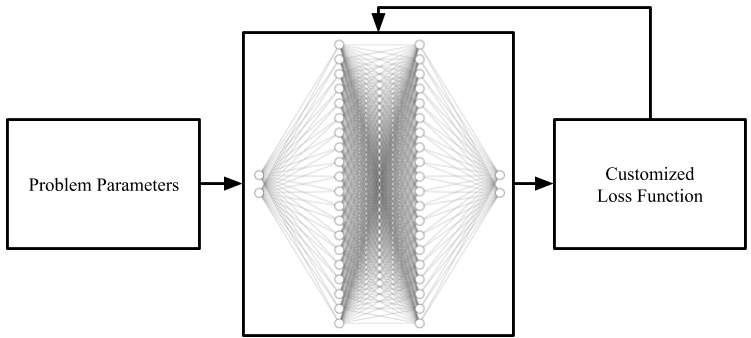}
    \caption{Structure of the unsupervised neural network for optimizing Rosenbrock's function.}
    \label{penalty}
\end{figure}

\begin{figure}[h]
\centering
  \includegraphics[width=1\linewidth]{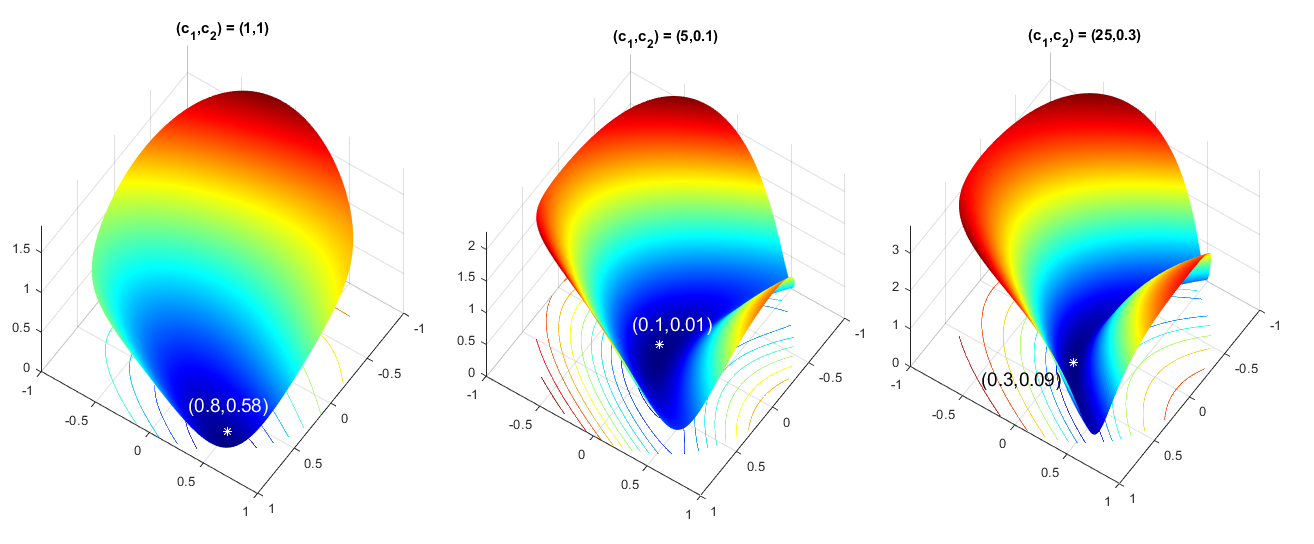}
  \caption{Rosenbrock function with different input parameters.}
\end{figure}
\begin{table}[!]
\begin{center}
\begin{footnotesize}
\caption{Results for Rosenbrock's function}
\label{ros_table}
\begin{tabular}{|c|c|c|c|}
  \hline
  Parameters & Interior Point method & DNN-generated solution \\ \hline
  $c_1=1,c_2=1$ &  $x_1=0.8082,x_2=0.5889$ &  $x_1=0.8394, x_2=0.6040$ \\
\hline
  $c_1=5,c_2=0.1$ &  $x_1=0.1000,x_2=0.0100$ &  $x_1=0.1014, x_2=0.0174$ \\
  \hline
  $c_1=25,c_2=0.3$ &  $x_1=0.3000,x_2=0.0900$ &  $x_1=0.3109, x_2=0.0957$ \\
\hline
\end{tabular}
\end{footnotesize}
\end{center}
\end{table}

%%%%%%%%%%%%%%%%%%%%%%%%%%%%%%%%%

\subsection{Rosenbrock's Function with Three Constraints}

\begin{equation}
\begin{cases}
    \min_x & c_1(x_2-x_1^2)^2 + (c_2-x_1)^2 \\
    \mathrm{s.t.} \ & x_1^2+x_2^2 \leq 1 \\
    \ & x_1 \leq -2.5 \\
    \ & x_2 \leq -1
\end{cases}
\end{equation}
We employed a DNN with a structure $(D_{in}, H_{layer1}, H_{layer2},H_{layer3},H_{layer4},H_{layer5}, D_{out}) = (2, 10,20,20,20,10, 2)$ and for training we used 1000 samples uniformly distributed in range $0\leq c_1 \leq 30$ and $0\leq c_2 \leq 1$. We used the ADAM optimizer for training the DNN and got the results shown below. %Pytorch code of this example is available in \href{https://colab.research.google.com/drive/1eLH_AYgUk1DWcqKYVdw24cq6Oiq-Ou3G?usp=sharing}{[Colab: Ros-3-constraints]}.

\begin{figure}[h]
\centering
  \includegraphics[width=.5\linewidth]{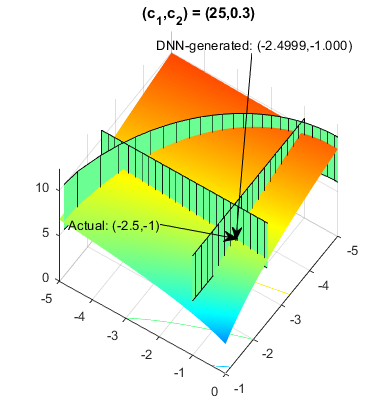}
  \caption{Rosenbrock function with three constraints}
\end{figure}
% \begin{figure}[h]
% \centering
%   \includegraphics[width=.6\linewidth]{loss_Rosenbrock_3_constraints.png}
%   \caption{Rosenbrock's test case with three constraints}
% \end{figure}

\subsection{Ackley's Function}
\begin{equation}
\begin{cases}
    \min_x -c_1 \exp(-c_2 \sqrt{c_3(x_1^2+x_2^2)}) - \exp(c_4(\cos(2\pi x_1)+\cos(2\pi x_2))) + e + c_5 \\
    \mathrm{s.t.} \ x_1^2+x_2^2 \leq 25
\end{cases}
\end{equation}
where, $c_1,\dots c_5$ are the parameters of this problem. We employed a DNN with a structure $(D_{in}, H_{layer1}, H_{layer2},H_{layer3},H_{layer4},H_{layer5}, D_{out}) = (5, 10,20,20,20,10, 2)$ and for training we used 1000 samples uniformly distributed in range $0\leq c_1,c_5 \leq 30$ and $0\leq c_2,c_3,c_4 \leq 1$. We used the ADAM optimizer for training the DNN and got the results shown in Table~\ref{ackley_table}. %Pytorch code of this example is available in \href{https://colab.research.google.com/drive/1ERzM7wEQAT2zMBa2Wtp7_h8p9EA6PbgV?usp=sharing}{[Colab: Ackley-1-constraint]}.
\begin{figure}[h]
\centering
  \includegraphics[width=1\linewidth]{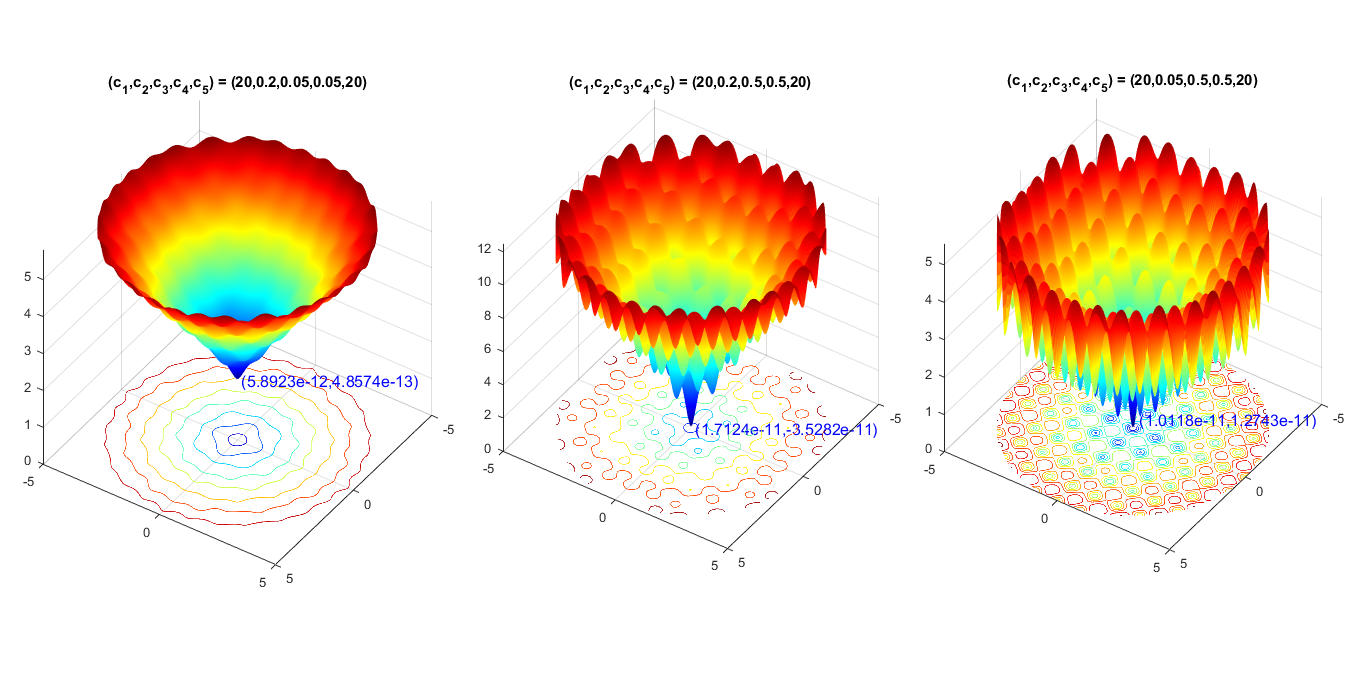}
  \caption{Ackley function}
\end{figure}

\begin{table}[h]
\begin{center}
\begin{footnotesize}
\caption{Results for Ackley's function}
\label{ackley_table}
\begin{tabular}{|p{4cm}|p{5cm}|p{4cm}|}
  \hline
  Parameters & Interior Point method & DNN-generated solution \\ \hline
  $c_1=20,c_2=0.2,c_3=0.05,c_4=0.05,c_5=20$ &  $x_1=5.8\times 10^{-12},x_2=12\times 10^{-13}$ &   $x_1 = -5.6177\times 10^{-6},  x_2 = 7.2256\times 10^{-5}$ \\
\hline
  $c_1=20,c_2=0.2,c_3=0.5,c_4=0.5,c_5=20$ &  $x_1=1.7\times 10^{-11},x_2=3.5\times 10^{-11}$ &  $x_1=-6.8992\times 10^{-6}, x_2 = 7.7887\times 10^{-5}$ \\
  \hline
  $c_1=20,c_2=0.05,c_3=0.5,c_4=0.5,c_5=20$ &  $x_1= 10^{-11},x_2=1.2\times 10^{-11}$ &  $x_1=-7.0035\times 10^{-6},  x_2=7.8982\times 10^{-5}$ \\
\hline
\end{tabular}
\end{footnotesize}
\end{center}
\end{table}

\subsection{Ackley's Function with Three Constraints}
\begin{equation}
\begin{cases}
    \min_x & -c_1 \exp(-c_2 \sqrt{c_3(x_1^2+x_2^2)}) - \exp(c_4(\cos(2\pi x_1)+\cos(2\pi x_2))) + e + c_5 \\
    \mathrm{s.t.} \ & x_1^2+x_2^2 \leq 1 \\
    \ & x_1 \leq -2.5 \\
    \ & x_2 \leq -1
\end{cases}
\end{equation}

We employed a DNN with a structure $(D_{in}, H_{layer1}, H_{layer2},H_{layer3},H_{layer4},H_{layer5}, D_{out}) = (5, 10,20,20,20,10, 2)$ and for training we used 1000 samples uniformly distributed in range $0\leq c_1,c_5 \leq 30$ and $0\leq c_2,c_3,c_4 \leq 1$. We used the ADAM optimizer for training the DNN and got the results shown below. % Pytorch code of this example is available in \href{https://colab.research.google.com/drive/1vd-Vg_kEiduP0vMa6zKt_wflgwfQn7gF?usp=sharing}{[Colab: Ackley-3-constraints]}.

\begin{figure}[h]
\centering
  \includegraphics[width=.5\linewidth]{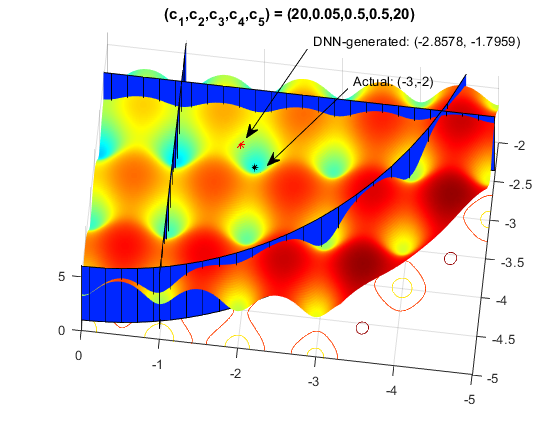}
  \caption{Ackley's function with three constraints}
\end{figure}

\section{Conclusion}
In this paper we proposed a DNN-based solution for solving constrained optimization problems. The novelty of our work is the piece-wise regularization method for imposing generic equality and inequality constraints. We tested our proposed method on famous artificial landscape objective functions under some nonlinear constraints and showed that with careful tuning and enough number of epochs we can achieve near-optimal feasible solutions with far less computational complexity.

\printbibliography

\end{document}